\documentclass{bmvc2k}
\usepackage{amsmath}
\title{Multiple Instance Curriculum Learning for Weakly Supervised Object Detection}

\addauthor{Siyang Li}{siyangl@usc.edu}{1}
\addauthor{Xiangxin Zhu}{xiangxin@google.com}{2}
\addauthor{Qin Huang}{qinhuang@usc.edu}{1}
\addauthor{Hao Xu}{hxu1@usc.edu}{1}
\addauthor{C.-C. Jay Kuo}{cckuo@sipi.usc.edu}{1}

\addinstitution{
	Department of Electrical Engineering\\
	University of Southern California\\
	Los Angeles, CA, USA
}
\addinstitution{
	Google, Inc.\\
	1600 Amphitheatre Pkwy\\
	Mountain View, CA, USA
}

\runninghead{S. Li et al.}{Multiple Instance Curriculum Learning}

\def\etal{\emph{et al}\bmvaOneDot}

\begin{document}

\maketitle
\begin{abstract}
	When supervising an object detector with weakly labeled data, most existing approaches are prone to trapping in the discriminative object parts, e.g., finding the face of a cat instead of the full body, due to lacking the supervision on the extent of full objects. To address this challenge, we incorporate object segmentation into the detector training, which guides the model to correctly localize the full objects. We propose the multiple instance curriculum learning (MICL) method, which injects curriculum learning (CL) into the multiple instance learning (MIL) framework. The MICL method starts by automatically picking the easy training examples, where the extent of the segmentation masks agree with detection bounding boxes. The training set is gradually expanded to include harder examples to train strong detectors that handle complex images. The proposed MICL method with segmentation in the loop outperforms the state-of-the-art weakly supervised object detectors by a substantial margin on the PASCAL VOC datasets.  
\end{abstract}

\section{Introduction}
Object detection is an important problem in computer vision. In recent years, a set of detectors based on convolutional neural networks (CNN) are proposed~\cite{FastRCNN2015, Faster2015,SSD2016}, which perform significantly better than traditional methods (e.g., \cite{DPM2010}). Those detectors need to be supervised with fully labeled data, where both object category and location (bounding boxes) are provided. However, we argue that such data are expensive in terms of labeling efforts and thus are not scalable. With the increase of dataset size, it becomes extremely difficult to label the locations of all object instances. 

In this work, we focus on object detection with weakly labeled data, where only image-level category labels are provided, while the object locations are unknown. This type of methods is attractive since image-level labels are usually much cheaper to obtain. For each image in a weakly labeled dataset, the image-level label tells both present and absent object categories. Thus, for each category, we have positive examples where at least one instance of that category is present as well as negative ones where no objects of that category exist. 

Some previous methods~\cite{Li_2016_CVPR, ContextLoc2016, Bilen_2016_CVPR} extract a set of object candidates via unsupervised object proposals (also called object candidates)~\cite{SelectiveSearch2013, EdgeBoxes2014, li2017box, carreira2010constrained} and then identify the best proposals that lead to high image classification scores for existing categories. However, the best proposals for image classification do not necessarily cover the full objects. (For example, to classify an image as ``cat'', seeing the face is already sufficient and even more robust than seeing the whole body, as the fluffy fur can be confused with other animals.) Specifically, the best proposals usually focus on discriminative object parts, which oftentimes do not overlap enough with the extent of full objects, and thus become false positives for detection.

%TODO briefly mention MIL, CL
%To reduce the false positives due to trapping in the discriminative parts, we propose the multiple instance curriculum learning (MICL) approach, which combines the commonly used multiple instance learning (MIL) framwork for weakly supervised object detection along with the easy-to-hard learning paradigm - curriculum learning~\cite{Curriculum2009}. The curriculum is designed based on a segmentation network, which is capable of expanding masks from the most discriminative regions. The workflow of the proposed MICL system is shown in Fig.~\ref{fig:overview}. First, a segmentation network and a detection network are trained seperately. To train the segmentation network, discriminative regions for existing object categories are automatically identified using \cite{CAM2016} and object seeds are generated as weak labels. To train the detection network, we identify the best object proposals for image classification. Afterwards, the curriculum learning methodology is adopted to update these two networks. Simple examples are identified by measuring consistency of results from the detection and the segmentation networks. They are used to re-train the detector and update object seeds so as to re-train the segmenter. This process is repeated, and the training set is gradually expanded from easy to hard examples. Once the training is finished, the detector is applied to test images directly given category-independent proposals from the Selective Search (SS)~\cite{SelectiveSearch2013}. 
%%%%%%%%%%%%%%%%%%%%%%%%%%%%%%%%%%%%%%%%%%%%%%%%%%%%%%%%%%%%%%%%
\begin{figure}[t]
	\centering
	\includegraphics[width=0.85\columnwidth]{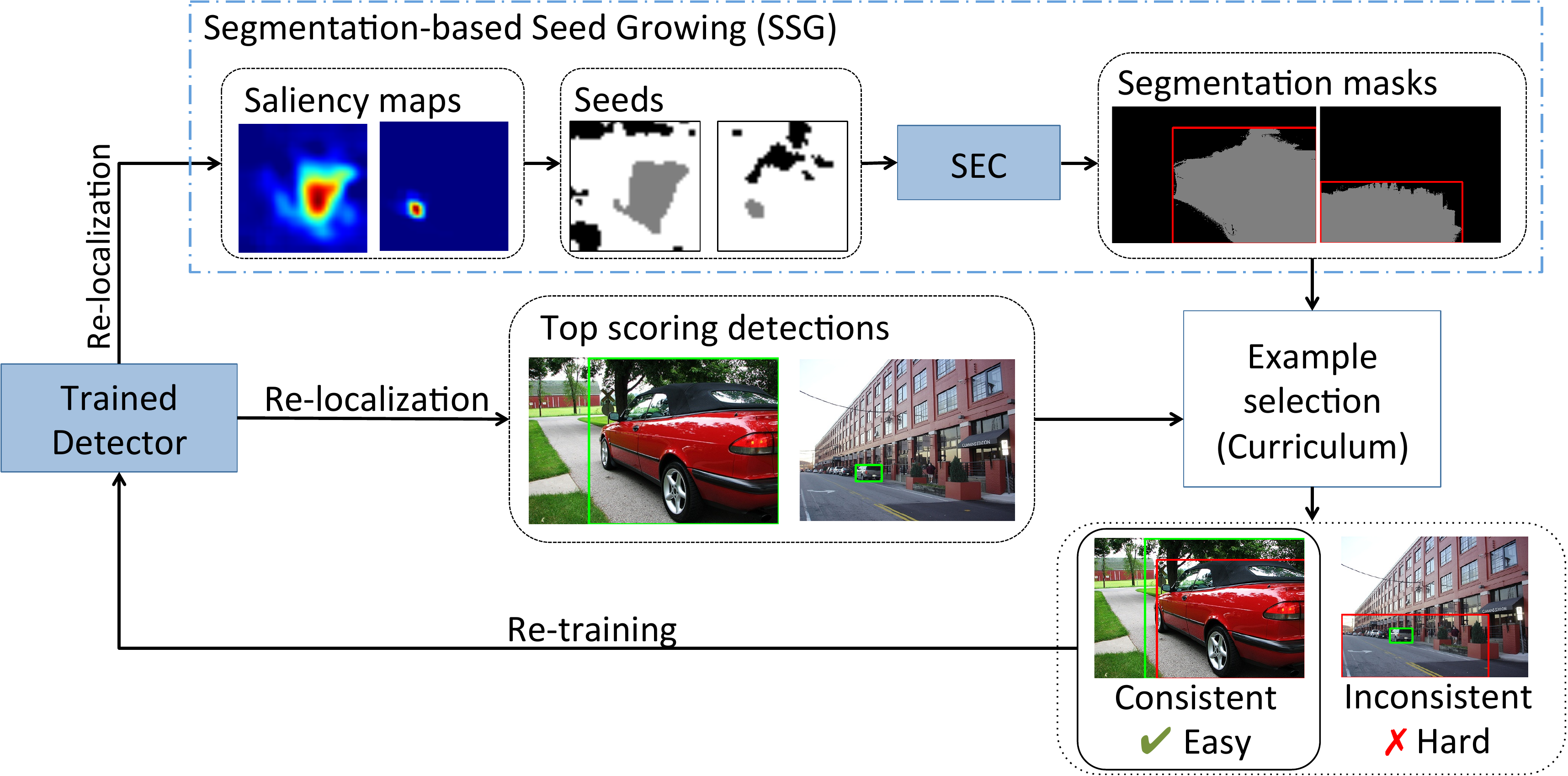}
	\caption{The training diagram of our MICL method. It iterates over re-localization and re-training. During re-localization, saliency maps are generated from the current detector. Segmentation seeds are obtained from the saliency maps, which later grow to segmentation masks.
		We use the segmentation masks to guide the detector to avoid being trapped in the discriminative parts. A curriculum is designed based on the segmentation masks and the current top scoring detections. With the curriculum, the multiple instance learning process can be organized in an easy-to-hard manner for the detector re-training.} \label{fig:overview} 
\end{figure}
%%%%%%%%%%%%%%%%%%%%%%%%%%%%%%%%%%%%%%%%%%%%%%%%%%%%%%%%%%%%%%%%
To reduce the false positives due to trapping in the discriminative parts, we use segmentation masks to guide the weakly supervised detector in the typical relocalization-and-retraining loop \cite{SizeEstimate2016,gokberk2014multi}. 
The segmentation process starts with a few seeds from the object saliency maps generated by the current detector. Then segmentation masks are obtained by expanding those seeds using the ``Seed, Expand and Constrain (SEC)'' method~\cite{SEC2016}.
%We only use the training examples where the outputs of the current detector agree with the segmentation masks to train the new detector. This way the new detector will learn to localize the whole objects which tend to be consistent to the segmentation, and will not be stuck at the discriminative parts.
One may use all generated masks to directly supervise the detector.	
However, it will be misled by the hard and noisy examples where the segmentation network fails to produce reasonably good object masks. To overcome this challenge, we propose the multiple instance curriculum learning (MICL) approach, which combines the commonly used multiple instance learning (MIL) framework with the easy-to-hard learning paradigm - curriculum learning (CL)~\cite{Curriculum2009}. The work flow of the proposed MICL system is shown in Fig.~\ref{fig:overview}. It learns from the ``easy'' examples only in the re-training step of MIL. The re-trained detector is later used to re-localize the segmentation seeds and object boxes. While this process iterates, the training set is gradually expanded from easy to hard examples so the detector learns to handle more complex examples. We identify the easiness of an example by examining the consistency between the results from the detector and the segmenter, without additional supervision on ``easiness'' required by traditional CL methods. Once the proposed MICL process is finished, the detector is applied to test images directly.

The contributions of this work are summarized as following.
First, we incorporate a semantic segmentation network to guide the detector to learn the extent of full objects and avoid being stuck at the discriminative object parts.
Second, we propose an MICL method by combining MIL with CL so that the detector is not misled by hard and unreliable examples, and our CL process does not require any additional supervision on the ``easiness'' of the training examples.
Third, we demonstrate the superior performance of the proposed MICL method as compared with the state-of-the-art weakly supervised detectors. 

\section{Related Work}\label{sec:related}
\textbf{Weakly supervised object detection.} The weakly supervised object detection problem is oftentimes treated as a multiple instance learning (MIL) task~\cite{bilen2014weakly,siva2011weakly,Song2014,song2014weakly,SizeEstimate2016, gokberk2014multi,bilen2015weakly,siva2012defence,siva2013looking}.  
Each image is considered as a bag of instances. An image is labeled as positive for one category if it contains at least one instance from that category, and an image is labeled as negative if no instance from that category appears in it.
As a non-convex optimization problem, MIL is sensitive to model initialization and many prior arts focus on good initialization strategies~\cite{Song2014, Deselaers2010}. Recently, people start to exploit the powerful CNN to solve the weakly supervised object detection problem. Oquab \etal~\cite{Oquab2015} convert the Alexnet~\cite{AlexNet2012} to a fully convolutional network (FCN) to obtain an object score map, but it gives only a rough location of objects. WSDNN~\cite{Bilen_2016_CVPR} constructs a two-branch detection network with a classification branch and a localization branch in parallel. 
ContextLoc~\cite{ContextLoc2016} is built on WSDNN and adds an additional context branch. DA~\cite{Li_2016_CVPR} proposes a domain adaptation approach to identify proposals corresponding to objects and uses them as pseudo location labels for detector training. Size Estimate (SE)~\cite{SizeEstimate2016} trains a separate size estimator with additional object size annotation and the detector is trained by feeding images with a decreasing estimated size. Singh \etal~\cite{TrackTransfer2016} take the advantage of videos and transfer the object appearance to images for object localization. 
Most of the aforementioned approaches suffer from trapping in discriminative regions as no specific supervision is provided to learn the extent of full objects. 

\noindent\textbf{Weakly supervised image semantic segmentation.} Similar with the location labels for detector training, the pixel-wise labels for semantic segmentation are also laborious to obtain. 
Consequently, many researchers focus on weakly supervised image semantic segmentation~\cite{Point2016,Scribblesup2016,SEC2016,papandreou2015weakly}, where only image-level labels or scribbles on objects are available for training. 
PointSup~\cite{Point2016} trains a segmentation network with one point annotation per category (or object) only and uses objectness prior to improve the results. Similarly, ScribbleSup~\cite{Scribblesup2016} trains the segmentation network with scribbles by alternating between the GraphCut~\cite{grabcut} algorithm and the FCN training~\cite{FCN2015}. SEC~\cite{SEC2016} generates coarse location cues from image classification networks as ``pseudo scribbles'', and constrains the training process with smoothness. In this work, SEC is used for segmentation network training with seeds.

\noindent\textbf{Curriculum learning.} The concept of curriculum
learning is proposed by Bengio \etal~\cite{Curriculum2009}, indicating that learning from easy to hard examples can be beneficial. It has been applied to various problems in computer vision~\cite{SizeEstimate2016,DiffEstimate2016,lee2011learningeasy,pentina2015curriculum}, where people present different definitions to ``easy'' examples. 
Some require human labelers to evaluate the difficulty level of images~\cite{pentina2015curriculum} while others measure the easiness based on labeling time~\cite{DiffEstimate2016}. In our work, we determine the easiness by measuring the consistency between a trained detector and the segmentation masks, without requiring additional human labeling.

\section{Proposed MICL Method}\label{sec:method}
The proposed MICL approach starts with initializing a detector to find the most salient candidate (Sec.~\ref{sec:init}). Meanwhile, we obtain the saliency maps from a trained classifier/detector. The saliency maps are thresholded to obtain segmentation seeds. Then a segmentation network is trained to grow object masks from those seeds (Sec.~\ref{sec:segmentation}).
After that, we inject the curriculum learning (CL) paradigm into the commonly used re-localization/re-training multiple instance learning (MIL) framework~\cite{gokberk2014multi} in the weakly supervised object detection problem, leading to the multiple instance curriculum learning (MICL) approach. With MICL we further train the detector to learn the extent of objects under the guidance of the segmentation network (Sec.~\ref{sec:iterative}).

\subsection{Detector Initialization}\label{sec:init}
\begin{figure}[hbtp]
	\centering
	\includegraphics[width=0.7\columnwidth]{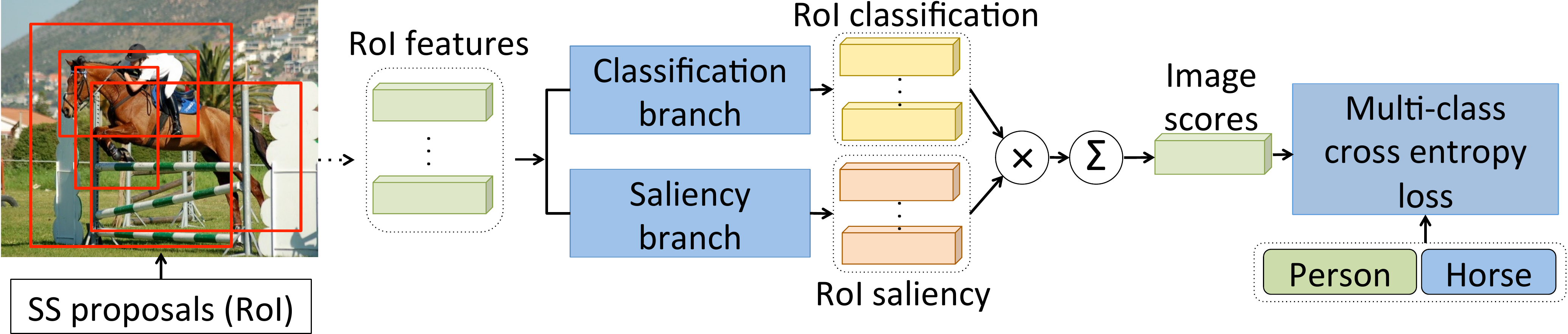}
	\caption{The detector with a saliency branch to find the most salient candidate (MSC) for image classification.} \label{fig:mhc_detector}
\end{figure}

To start the MIL process detailed in Sec.~\ref{sec:iterative}, we need an initial detector with certain localization capability. We achieve this by first training a whole image classifier and then tuning it into a detector to identify the most salient candidate. Similar with the methods described in \cite{Bilen_2016_CVPR,ContextLoc2016,Attention2016}, we first extract SS~\cite{SelectiveSearch2013} object proposals. In order to find the most salient candidate (MSC) among the proposals, we then add a ``saliency'' branch composed of an FC layer in parallel with the classification branch, as shown in Fig.~\ref{fig:mhc_detector}.
%%%%%%%%%%%%%%%%%%%%%%%%%%%%%%%%%%%%%%%%%%%%%%%%%%%%%%%%%%%%%%%%
%%%%%%%%%%%%%%%%%%%%%%%%%%%%%%%%%%%%%%%%%%%%%%%%%%%%%%%%%%%%%%%%
This branch takes the region of interest (RoI, denoted by $r$) features from the second last layer of the VGG16 network~\cite{VGG2014} as the input, and computes its saliency. Then the softmax operation is applied, so that the saliency scores of all RoIs for category $c$, denoted by $h(c; r)$, sum to one, i.e. $\sum_{r} h(c;r)= 1$. The saliency score is used to aggregate the RoI classification score $p(c;r)$ into image-level scores:
\begin{align}
p(c) = \sum_{r} h(c; r) p(c; r)/\sum_{r} h(c; r) = \sum_{r} h(c; r) p(c; r).
\end{align}

Then, the whole network can be trained with image-level
labels using the multi-label cross entropy loss. 
We rank the RoIs by the combined scores $h(c; r) p(c; r)$ and record the top scoring RoI for the curriculum learning process detailed in Sec.~\ref{sec:iterative}. 

\subsection{Segmentation-based Seed Growing (SSG)}\label{sec:segmentation}
In this module, we first obtain saliency maps from a classification (a single region of interest) or detection (multiple regions of interest) network and then train a segmentation network to expand object masks from the salient regions.

\noindent\textbf{Saliency map for a single region.}
Several methods~\cite{CAM2016,Grad2013,ExciteBack2016} are
proposed to identify the saliency map from a trained image classification network automatically.
The saliency map for category $c$, denoted by $A(x, y; c)$, describes the discriminative power of the cell, $(x, y)$, on the feature map $f(x, y, k)$ generated by the trained classification network. Here, we use CAM~\cite{CAM2016} to obtain the saliency maps for existing categories and Grad~\cite{Grad2013} for background as elaborated below. 

Being applied to a GAP classification network\footnote{GAP refers to global average pooling. GAP networks refer to the architecture with only the top layer being fully connected (FC). Prior to the FC layer is a GAP layer which averages the 3-D feature map into a feature vector.}, the CAM saliency map is defined as
\begin{align}\label{eq:cam}
A^{CAM}(x, y;c) = \sum_{k}f(x, y, k)w(k;c),
\end{align}

\noindent where $(x, y)$ is the cell coordinates of the feature map from the last convolutional layer, $f(x,y,k)$ is the response at the $k$-th unit at this layer and $w(k;c)$ is the weight parameters of the fully connected (FC) layer corresponding to category $c$.

The Grad background saliency maps are defined as
\begin{align}\label{eq:grad}
A^{Grad}_{BG}(x, y) = 1 - \max_{c\in C^{+}}\frac{1}{Z(c)}\max_{k}\left 
|\frac{\partial p(c)}{\partial f(x,y,k)}\right|,
\end{align}
where $p(c)$ is the output of the classification network,
indicating the probability that the input image belongs to
category $c$, and $Z(c)$ is a normalization factor so that
the maximum saliency for all existing categories, denoted by $C^{+}$,  is normalized to one. More specifically,
\begin{align}
Z(c) = \max_{x, y, k}\left 
|\frac{\partial p(c)}{\partial f(x,y,k)}\right|.
\end{align}

\begin{figure}[h]
	\centering
	\includegraphics[width=0.46\columnwidth]{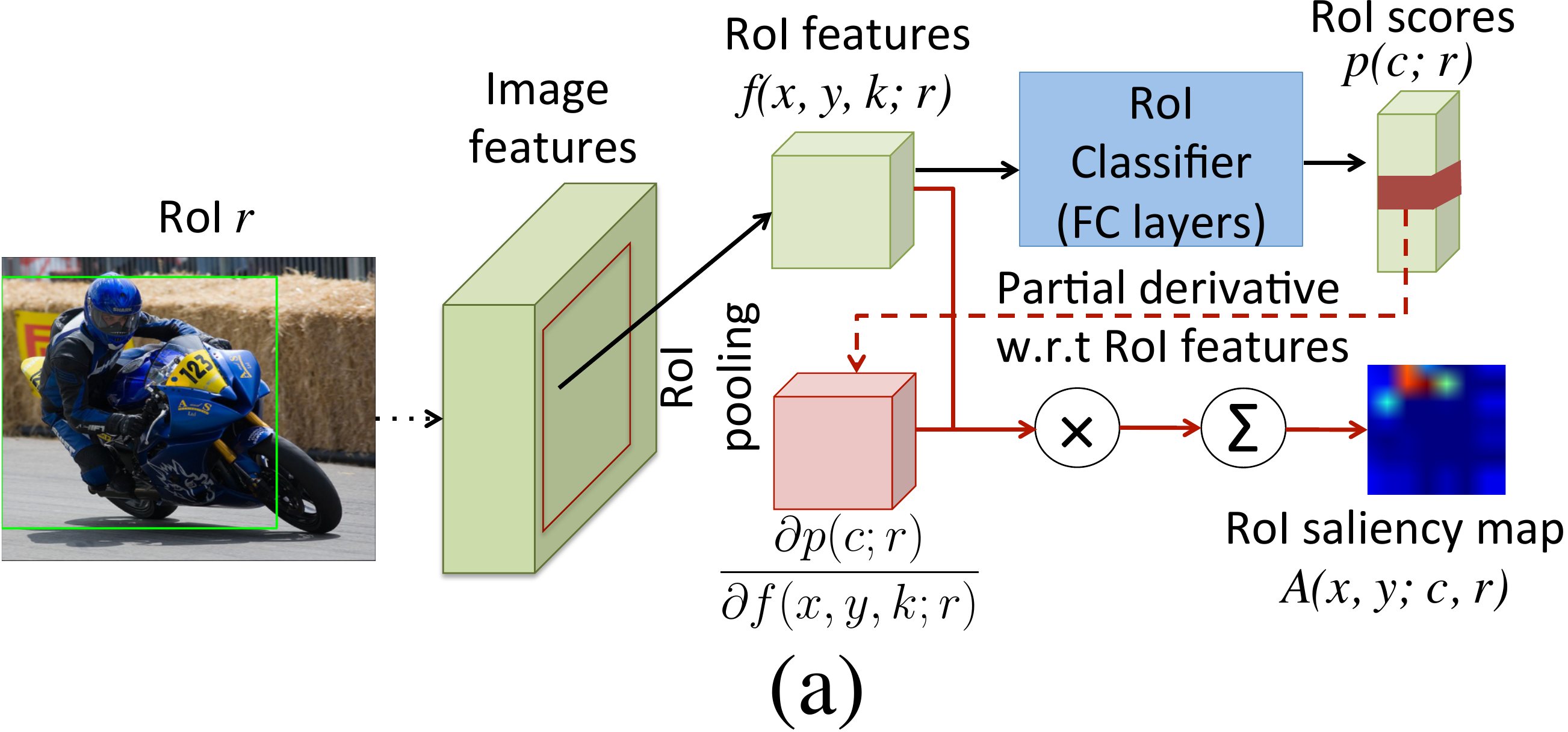}
	\hspace{0.01\columnwidth}
	\includegraphics[width=0.48\columnwidth]{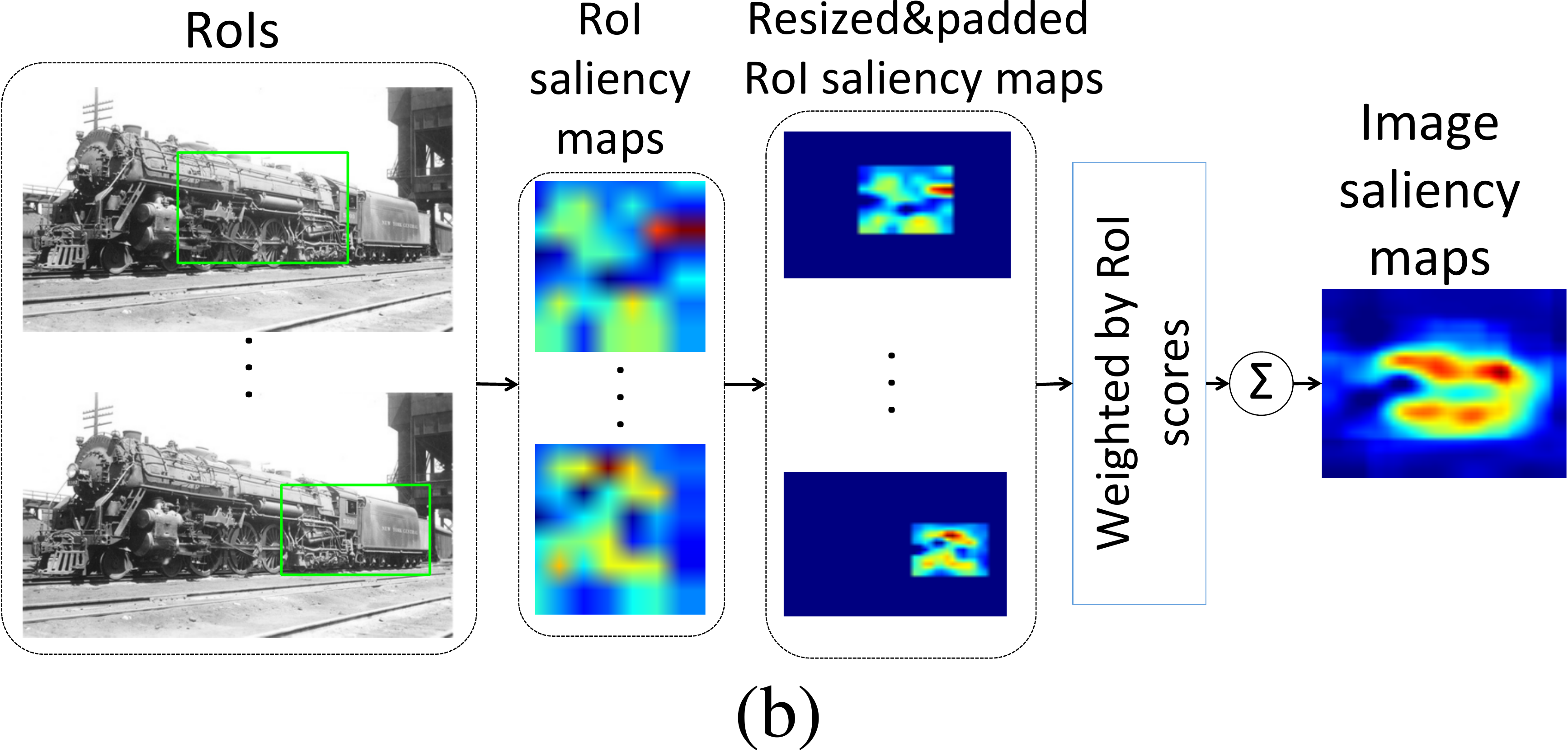}
	\caption{(a) Compute the saliency map in a RoI, and (b) aggregate 
		RoI saliency maps to the image saliency maps.}\label{fig:seed_update}
\end{figure}

\noindent\textbf{Aggregated saliency map from multiple regions.} The outputs of a detector are essentially a group
of bounding boxes with classification scores. We propose a
generalized saliency map mechanism that can be applied to
detectors with the RoI pooling layer (e.g., Fast R-CNN~\cite{FastRCNN2015} as used in this work). As illustrated in Fig.~\ref{fig:seed_update}\textcolor{red}{(a)}, given a RoI $r$ and its feature map, denoted by $f(x,y,k;r)$ (which has a fixed size because the classifier is composed of FC layers), obtained by the RoI pooling operator, one can compute the saliency map within $r$ as
\begin{align}\label{eq:generalized_cam}
A(x, y; c, r) = \sum_{k}\frac{\partial p(c; r)}{\partial f(x,y,k; r)} 
\otimes f(x, y, k;r),
\end{align}
where $\otimes$ denotes element-wise multiplication and $p(c; r)$ is the classification score for $r$. To obtain the saliency maps for the entire image, the RoI saliency maps are aggregated via
\begin{align}\label{eq:roi_attn_add}
%TODO
A(x, y; c) = \sum_{r=1}^{N}\hat{A}(x,y; c, r) p(c; r),
\end{align}
where $\hat{A}(x,y; c, r)$ is obtained by resizing $A(x, y; c, r)$ to the RoI size using bilinear interpolation and then padding it to the image size, as shown in Fig.~\ref{fig:seed_update}\textcolor{red}{(b)}. 

It is worthwhile to point out that Eq.~(\ref{eq:generalized_cam}) is reduced to the single region or the whole image saliency map for a classification network with no RoI inputs (i.e., the entire image as one RoI). When the image classification network is a GAP network, it can be further reduced to Eq.~(\ref{eq:cam}) as derived in the original CAM~\cite{CAM2016}. These relations are explained in the supplementary materials.
%%%%%%%%%%%%%%%%%%%%%%%%%%%%%%%%%%%%%%%%%%%%%%%%%%%%%%%%%%%%%%%%
%%%%%%%%%%%%%%%%%%%%%%%%%%%%%%%%%%%%%%%%%%%%%%%%%%%%%%%%%%%%%%%%

\noindent\textbf{Object mask growing from discriminative regions.}
Segmentation seeds are obtained by thresholding these
saliency maps and the seeds from CAM and Grad are simply pooled together. We adopt the SEC method~\cite{SEC2016} to train the DeepLab~\cite{DeepLab2014} network to expand masks from those seeds. In the first round of training, seeds come from a classification network while in later rounds they are from the currently trained detector.
The trained segmentation network is applied to all training images and a bounding box is drawn around the largest connected component in the mask. Thus one instance from each existing category is localized. The location information will guide the detector training process described in Sec.~\ref{sec:iterative}.
	
\subsection{Multiple Instance Curriculum Learning}\label{sec:iterative}
The commonly used MIL framework usually starts with an initial detector and then alternates between updating the boxes (re-localization) and updating the model (re-training). In re-locolization, the current detector is applied to training images and the highest-scoring box is saved. In re-training, the detector is re-trained on the saved boxes. 

To re-train the initialized detector from Sec.~\ref{sec:init}, one may use the highest-score boxes produced by itself, but it get stuck easily at the same box. Alternatively, on can use the bounding boxes from the SSG network, but those boxes may not be reliable due to inaccurate segmentation seeds. In other words, relying solely on the initial detector or the segmenter leads to sub-optimal results. To avoid misleading the detector by unreliable boxes, we guide the detector by organizing the MIL process on a curriculum that requires the detection boxes and segmentation masks to agree with each other. Details are elaborated below.

\noindent\textbf{SSG-Guided Detector Training.} 
The easy-to-hard learning principle proposed in \cite{Curriculum2009} has been proved helpful in training a weakly supervised object detector~\cite{SizeEstimate2016,DiffEstimate2016}. However, most previous methods require additional human supervision or additional object size information to determine the hardness of a training example, which is expensive to acquire. Instead of seeking additional supervision, we determine the hardness of an example by measuring the consistency between the outputs from the detector and the SSG network. The consistency is defined as the intersection over union (IoU) of the boxes from the SSG and the detector:
\begin{equation}
S(c; z)= IoU(B^{DET}(c; z), B^{SSG}(c; z)),
\end{equation}
where $z$ represents a positive example for category $c$;
$B^{DET}(c; z)$ and $B^{SSG}(c; z)$ stand for the predicted bounding boxes of the object by the detector and the SSG network, respectively. As shown in Fig.~\ref{fig:overview}, an example is considered \textit{easy} if
\begin{align}\label{eq:agreement}
S(c; z) \geq T,
\end{align}
where $T$ is a threshold to control the hardness of the selected examples.
We argue the validity of this criterion from two perspectives. First,
those examples are \textit{easier} because the
goal for the detector is to mimic the mask expansion ability
of SSG, and one example would appear easier if the detector
already produces something similar (i.e., the
gap between achieved results and the learning target is
small). Second, the object localization on those examples
are confirmed by both the detector and SSG, meaning that those predicted locations are more reliable. In other words, if
$B^{SSG}(c; z)$ significantly deviates from $B^{DET}(c; z)$,
it tends to be unreliable. We verify the reliability of $B^{SSG}(c; z)$ on the selected examples in Sec.~\ref{sec:analysis}.

For an existing category $c$, one instance is localized by taking the average of $B^{DET}(c;z)$ and $B^{SSG}(c;z)$ on the selected examples. Those localized instances on easy examples are used for further detector training in a fully supervised manner. In this work, we use the popular Fast R-CNN detector~\cite{FastRCNN2015} with Selective Search~\cite{SelectiveSearch2013} as the object proposal generator.

\noindent\textbf{Re-localization and re-training.} The detector trained on easy examples lacks the ability to handle hard examples because it focuses on easy ones in the aforementioned training round. Thus, we gradually include more training examples by adopting the re-localization/re-training iterations in the MIL framework~\cite{gokberk2014multi}, illustrated in Fig.~\ref{fig:overview}. In the re-localization step, the trained detector is applied to the whole training set and the highest scoring boxes for existing categories are recorded as the new $B^{DET}(c;z)$. Meanwhile, the outputs from the detector are used to re-localize segmentation seeds, based on which the SSG network is re-trained to generate new $B^{SSG}(c;z)$. By applying the same example selection criterion, another training subset, containing more examples because their results are more similar after learning from each other, is identified and the Fast R-CNN detector is re-trained. The MIL process alternates between re-training and re-localization until all training examples are included.
After training is finished, the detector is applied directly to test images.

\section{Experiments}\label{sec:experiment}

\subsection{Experiment Settings}
\noindent\textbf{Datasets.} To evaluate the performance of the weakly supervised MICL detector, we conduct experiments on the PASCAL VOC 2007 and 2012 datasets (abbreviated as VOC07 and VOC12 below)~\cite{Pascal2010}, where 20 object categories are labeled. For the MICL detector training, we only use image-level labels, with no human labeled bounding boxes involved.

\noindent\textbf{Evaluation metrics.} Following the evaluation of fully supervised detectors, we use the average precision (AP) for each category on the test set as the performance metric. Besides, we choose another metric, the Correct Location (CorLoc)~\cite{Corloc2012}, to evaluate weakly supervised detectors, which is usually applied to training images. The CorLoc is the percentage of the true positives among the most confident predicted boxes for existing categories. A predicted box is a true positive if it overlaps sufficiently with one of the ground truth object boxes. The IoU threshold is set to 50\% for both metrics.

\noindent\textbf{Implementation Details.}
The backbone architecture for all modules is the VGG16~\cite{VGG2014} network, which is pre-trained on the ImageNet dataset~\cite{ImageNet2009}. The whole system is implemented with the TensorFlow\cite{Tensorflow2016} library. The saliency map thresholds in SSG are 0.2 for objects and 0.9 for background\footnote{We reduce the background threshold to make background seeds take at least 10\% of saliency map, as in \cite{SEC2016}.}. The parameter $T$ in Eq.~\ref{eq:agreement} balances the number of selected examples and their easiness. Higher T means easier but fewer examples. It trades off between ``overfitting to fewer clean examples'' and ``learning from large noisy data''. We find empirically that $T=0.5$ gives good results. 
Other details are found in the supplementary materials.

\subsection{Experimental Results}
%%%%%%%%%%%%%%%%%%%%%%%%%%%%%%%%%%%%%%%%%%%%%%%%%%%%%%%%%%%%%%%%
\begin{figure}
	\centering
	\includegraphics[height=0.14\columnwidth]{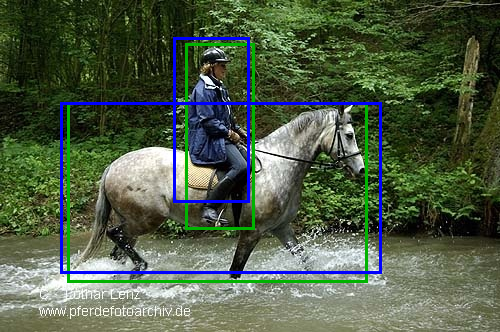}
	\includegraphics[height=0.14\columnwidth]{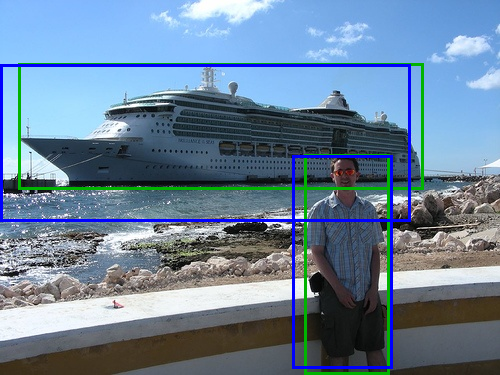}
	\includegraphics[height=0.14\columnwidth]{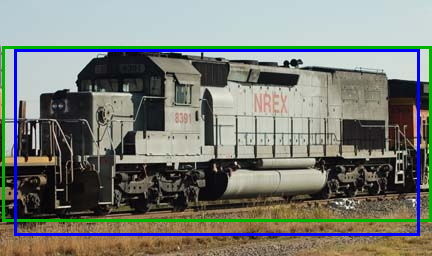}
	\includegraphics[height=0.14\columnwidth]{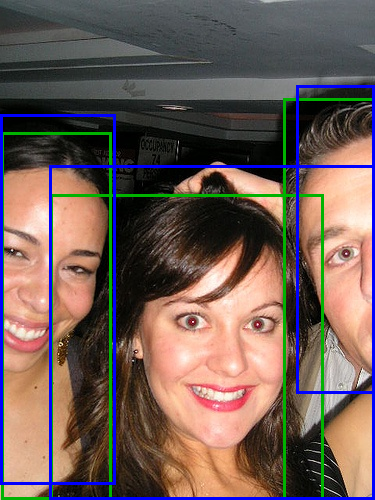}
	\includegraphics[height=0.14\columnwidth]{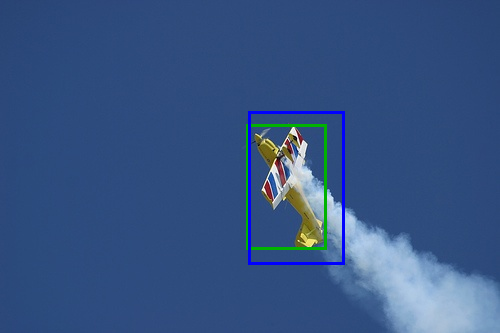}
	\includegraphics[height=0.13\columnwidth]{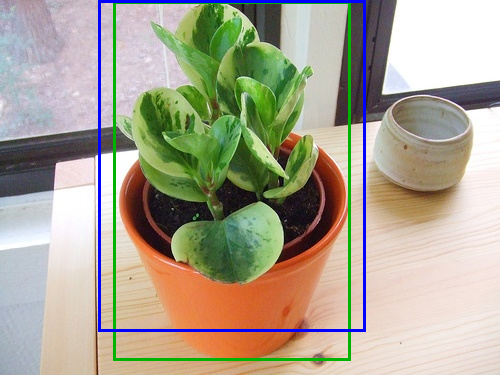}
	\includegraphics[height=0.13\columnwidth]{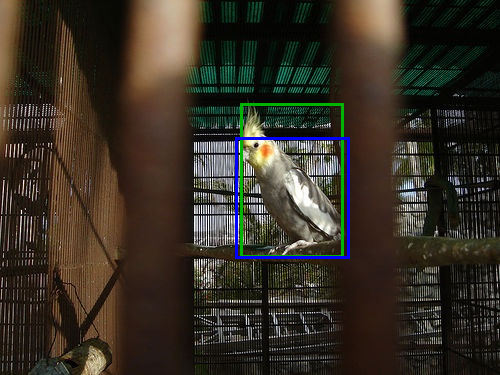}
	\includegraphics[height=0.13\columnwidth]{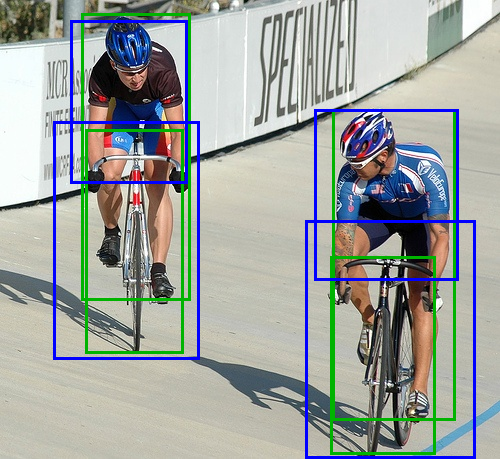}
	\includegraphics[height=0.13\columnwidth]{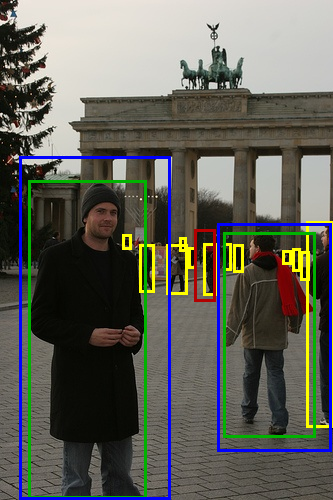}	
	\includegraphics[height=0.13\columnwidth]{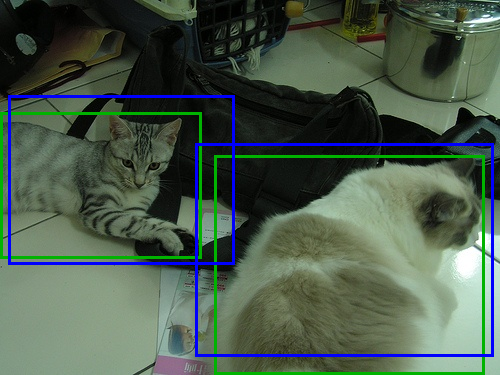}
	\includegraphics[height=0.13\columnwidth]{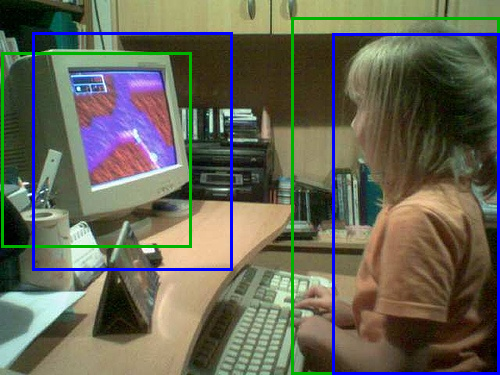}
	\includegraphics[height=0.14\columnwidth]{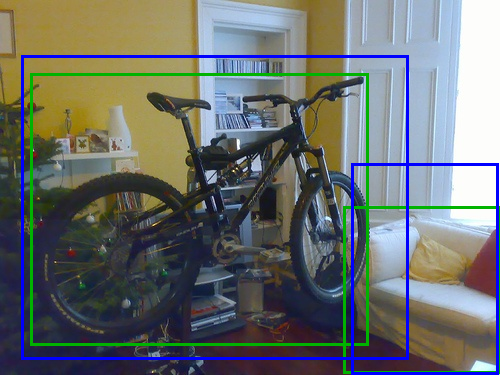}
	\includegraphics[height=0.14\columnwidth]{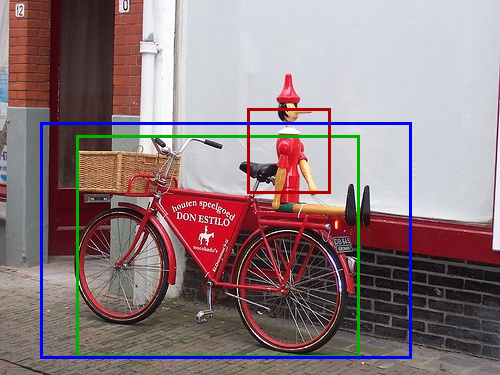}
	\includegraphics[height=0.14\columnwidth]{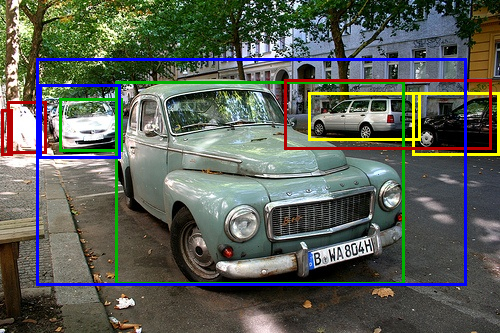}
	\includegraphics[height=0.14\columnwidth]{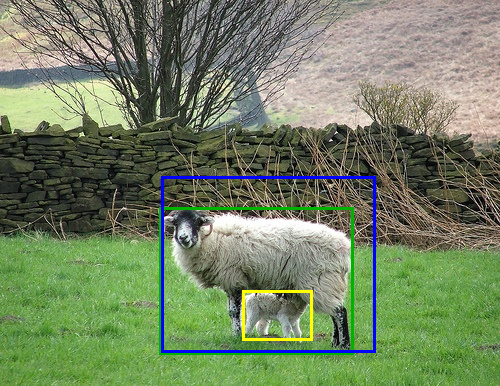}
	\includegraphics[height=0.14\columnwidth]{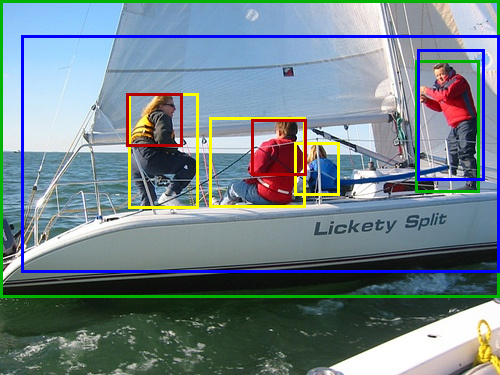}
	\caption{Qualitative detection results, where the correctly detected ground truth objects are in green boxes and blue boxes represent correspondingly predicted locations. Objects that the model fails to detect are in yellow boxes and false positive detections are in red.}\label{fig:result_example}
\end{figure}
%%%%%%%%%%%%%%%%%%%%%%%%%%%%%%%%%%%%%%%%%%%%%%%%%%%%%%%%%%%%%%%%
%%%%%%%%%%%%%%%%%%%%%%%%%%%%%%%%%%%%%%%%%%%%%%%%%%%%%%%%%%%%%%%%
\begin{table}[btp]
	\fontsize{7}{8}\selectfont
	\begin{center}
		\setlength\tabcolsep{1pt} 
		\begin{tabular}{|l|cccccccccccccccccccc|c|}
			\hline
			Method & aero &  bike & bird & boat & bottle & bus & car & cat & chair & cow & table & dog & horse & mbike & prsn &plant &sheep &sofa &train &tv & mAP\\
			\hline\hline
			WSDNN-Ens\cite{Bilen_2016_CVPR} &46.4 &\textbf{58.3} &35.5 &25.9 &14.0 &\textbf{66.7} &53.0 &39.2 &8.9 &41.8 &26.6 &38.6 &44.7 &59.0 &10.8 &17.3 &\textbf{40.7} &49.6 &\textbf{56.9} &\textbf{50.8} &39.3\\
			WSDNN\cite{Bilen_2016_CVPR} &43.6 &50.4 &32.2 &26.0 &9.8 &58.5 &50.4 &30.9 &7.9 &36.1 &18.2 &31.7 &41.4 &52.6 &8.8 &14.0 &37.8 &46.9 &53.4 &47.9 &34.9\\
			DA\cite{Li_2016_CVPR} &54.5 &47.4 &41.3 &20.8 &\textbf{17.7} &51.9 &\textbf{63.5} &46.1 &\textbf{21.8} &\textbf{57.1} &22.1 &34.4 &50.5 &\textbf{61.8} &16.2 &\textbf{29.9} &\textbf{40.7} &15.9 &55.3 &40.2 &39.5 \\
			ConLoc\cite{ContextLoc2016} &57.1 &52.0 &31.5 &7.6 &11.5 &55.0 &53.1 &34.1 &1.7 &33.1 &49.2 &42.0 &47.3 &56.6 &15.3 &12.8 &24.8 &48.9 &44.4 &47.8 &36.3  \\
			Attn\cite{Attention2016} &48.8 &45.9 &37.4 &\textbf{26.9} &9.2 &50.7 &43.4 &43.6 &10.6 &35.9 &27.0 &38.6 &48.5 &43.8 &24.7 &12.1 &29.0 &23.2 &48.8 &41.9 &34.5\\
			SE\cite{SizeEstimate2016} &- &- &- &- &- &- &- &- &- &- &- &- &- &- &- &- &- &- &- &-&37.2 \\
			\hline
			MICL &\textbf{61.2} &51.9 &\textbf{47.1} &13.5 &10.1 &52.1 &56.9 &\textbf{71.0} &7.6 &36.4 &\textbf{49.7} &\textbf{64.5} &\textbf{63.0} &57.8 &\textbf{27.9} &16.6 &30.4 &\textbf{53.8} &41.1 &40.3 &\textbf{42.6}\\
			\hline
		\end{tabular}
	\end{center}
	\caption{Comparison of mAP values on the VOC07
		\textit{test} set. Note that the gaps between
		the previous approaches and our method are particularly large
		on categories such as {\bf``cat''}, {\bf``dog''}
		and {\bf``horse''}. The improvements are from the SSG network, which grows the bounding box to cover the full objects from the discriminative parts (i.e., faces).}
	\label{tbl:map}
\end{table}
%%%%%%%%%%%%%%%%%%%%%%%%%%%%%%%%%%%%%%%%%%%%%%%%%%%%%%%%%%%%%%%%
The average precision (AP) on the VOC07 \textit{test} set is shown in Tab.~\ref{tbl:map}. An mAP of 42.6\% is achieved by our MICL method, with 3.1\% higher than the second best method, DA~\cite{Li_2016_CVPR}. Note that DA~\cite{Li_2016_CVPR} needs to cache the RoI features from pretrained CNN models for MIL and demands highly on disk space. Our proposed method avoids feature caching and thus is more scalable. The third best method is from ensembles~\cite{Bilen_2016_CVPR} (WSDNN-Ens in Tab.~\ref{tbl:map}). If compared directly to the results without ensembles (WSDNN in Tab.~\ref{tbl:map}), our method is 7.7\% superior. Some visualized detection results are shown in Fig.~\ref{fig:result_example}. 

Tab.~\ref{tbl:corloc} shows the CorLoc evaluation on the VOC07 \textit{trainval} set. We achieve 2.5\% higher in the CorLoc than WSDNN-Ens~\cite{Bilen_2016_CVPR}. Also, as compared with that of DA~\cite{Li_2016_CVPR}, which ranks the second in the mAP, our MICL method is superior by 8.1\%. Note that in terms of both AP and CorLoc, our MICL detector performs much better on certain categories such as ``cat'', ``dog'' and ``horse''. Objects in these categories usually have very discriminative parts (i.e. faces). Improvements on those categories are from the SSG network, which grows the bounding boxes from the discriminative regions. 

%%%%%%%%%%%%%%%%%%%%%%%%%%%%%%%%%%%%%%%%%%%%%%%%%%%%%%%%%%%%%%%%
\begin{table}[h]
	\fontsize{7}{8}\selectfont
	\begin{center}
		\setlength\tabcolsep{1pt} 
		\begin{tabular}{|l|cccccccccccccccccccc|c|}
			\hline
			Method & aero &  bike & bird & boat & bottle & bus & car & cat & chair & cow & table & dog & horse & mbike & prsn &plant &sheep &sofa &train &tv & Avg.\\
			\hline\hline
			WSDNN-Ens\cite{Bilen_2016_CVPR} &68.9 &\textbf{68.7} &65.2 &42.5 &\textbf{40.6} &72.6 &75.2 &53.7 &29.7 &68.1 &33.5 &45.6 &65.9 &\textbf{86.1} &27.5 &44.9 &76.0 &62.4 &66.3 &\textbf{66.8} &58.0\\
			WSDNN\cite{Bilen_2016_CVPR} &65.1 &63.4 &59.7 &\textbf{45.9} &38.5 &69.4 &77.0 &50.7 &30.1 &68.8 &34.0 &37.3 &61.0 &82.9 &25.1 &42.9 &\textbf{79.2} &59.4 &\textbf{68.2} &64.1 &56.1 \\
			DA\cite{Li_2016_CVPR} &78.2 &67.1 &61.8 &38.1 &36.1 &61.8 &\textbf{78.8} &55.2 &28.5 &68.8 &18.5 &49.2 &64.1 &73.5 &21.4 &47.4 &64.6 &22.3 &60.9 &52.3 &52.4 \\
			ConLoc\cite{ContextLoc2016} &83.3 &68.6 &54.7 &23.4 &18.3 &\textbf{73.6} &74.1 &54.1 &8.6 &65.1 &47.1 &59.5 &67.0 &83.5 &35.3 &39.9 &67.0 &49.7 &63.5 &65.2 &55.1 \\
			WSC\cite{Cascade2016} &83.9 &72.8 &64.5 &44.1 &40.1 &65.7 &82.5 &58.9 &\textbf{33.7} &\textbf{72.5} &25.6 &53.7 &67.4 &77.4 &26.8 &\textbf{49.1} &68.1 &27.9 &64.5 &55.7 &56.7 \\
			\hline
			MICL &\textbf{85.3} &58.4 &\textbf{68.5} &30.4 &20.9 &67.2 &77.1 &\textbf{84.6} &24.3 &69.5 &\textbf{51.5} &\textbf{80.0} &\textbf{79.8}  &85.3  &\textbf{46.2}  &44.9  &52.1 &\textbf{64.6} &61.3  &60.0 &\textbf{60.5}\\
			\hline
		\end{tabular}
	\end{center}
	\caption{Comparison of the CorLoc on the VOC07
		\textit{trainval} set.}
	%\todo{same in this table, boldface best numbers, and have the baseline names.}
	%TODO(Need to update numbers)
	\label{tbl:corloc}
\end{table}
%%%%%%%%%%%%%%%%%%%%%%%%%%%%%%%%%%%%%%%%%%%%%%%%%%%%%%%%%%%%%%%%

To our best knowledge, only \cite{ContextLoc2016} and
\cite{Li_2016_CVPR} reported results on the VOC12 dataset. We follow the training/testing split in \cite{ContextLoc2016} and show the results in Table~\ref{tbl:voc2012}. We improve the mAP and the CorLoc by 3.6\% and 8.4\%, respectively. For performance benchmarking with \cite{Li_2016_CVPR}, we estimate the mAP of our MICL method on the VOC12 \textit{val} set by applying the detector trained on the VOC07 \textit{trainval} set. The obtained mAP is 37.8\%, which is 8\% higher than \cite{Li_2016_CVPR}. We should point out that the VOC07 \textit{trainval} set does not overlap with the VOC12 \textit{val} set and its size is about the same as the VOC12 \textit{train} set used in \cite{Li_2016_CVPR}. 

%%%%%%%%%%%%%%%%%%%%%%%%%%%%%%%%%%%%%%%%%%%%%%%%%%%%%%%%%%%%%%%%
\begin{table}[h]
	\begin{center}
		\fontsize{6.8}{7.5}\selectfont
		\setlength\tabcolsep{1pt} 
		\begin{tabular}{|l|l|cccccccccccccccccccc|c|}
			\hline
			&Method & aero &  bike & bird & boat & bottle & bus & car & cat & chair & cow & table & dog & horse & mbike & prsn &plant &sheep &sofa &train &tv & Avg.\\
			\hline\hline
			AP&ConLoc\cite{ContextLoc2016} &64.0 &54.9 &36.4 &\textbf{8.1}&\textbf{12.6} &\textbf{53.}1 &40.5 &28.4 &6.6 &\textbf{35.3} &\textbf{34.4} &49.1 &42.6 &62.4 &\textbf{19.8} &15.2 &\textbf{27.0} &33.1 &\textbf{33.0} &\textbf{50.0} &35.3\\
			&MICL &\textbf{65.5} &\textbf{57.3} &\textbf{53.4} &5.4 &11.5 &48.8 &\textbf{45.4} &\textbf{80.5} &\textbf{7.6} &35.2 &25.3 &\textbf{75.8} &\textbf{59.5} &\textbf{68.8} &18.0 &\textbf{17.0} &24.7 &\textbf{37.7} &25.8 &14.1 &\textbf{38.9}\\
			\hline\hline
			Cor-&ConLoc\cite{ContextLoc2016} &78.3 &70.8 &52.5 &\textbf{34.7} &\textbf{36.6} &\textbf{80.0} &58.7 &38.6 &27.7 &\textbf{71.2} &32.3 &48.7 &76.2 &77.4 &16.0 &48.4 &\textbf{69.9} &47.5 &\textbf{66.9} &\textbf{62.9} &54.8\\
			Loc&MICL &\textbf{84.9} &\textbf{78.6} &\textbf{76.9} &30.1 &32.6 &\textbf{80.0} &\textbf{69.7} &\textbf{90.6} &\textbf{32.1} & 67.7 &\textbf{47.4} &\textbf{85.5} &\textbf{85.3} &\textbf{85.9} &\textbf{41.4} &\textbf{50.7} &62.8 &\textbf{62.7} &57.9 &41.7 &\textbf{63.2} \\
			\hline
		\end{tabular}
	\end{center}
	\caption{Results on the VOC12 dataset, where the AP and the CorLoc are measured on
		the \textit{test} and \textit{trainval} set, respectively.}
	\label{tbl:voc2012}
\end{table}
%%%%%%%%%%%%%%%%%%%%%%%%%%%%%%%%%%%%%%%%%%%%%%%%%%%%%%%%%%%%%%%%
\vspace{-0.5cm}
\subsection{Analyses}\label{sec:analysis}
\begin{figure}[h]
	\centering
	\begin{minipage}{0.48\columnwidth}
		\centering
		\includegraphics[height=0.4\columnwidth]{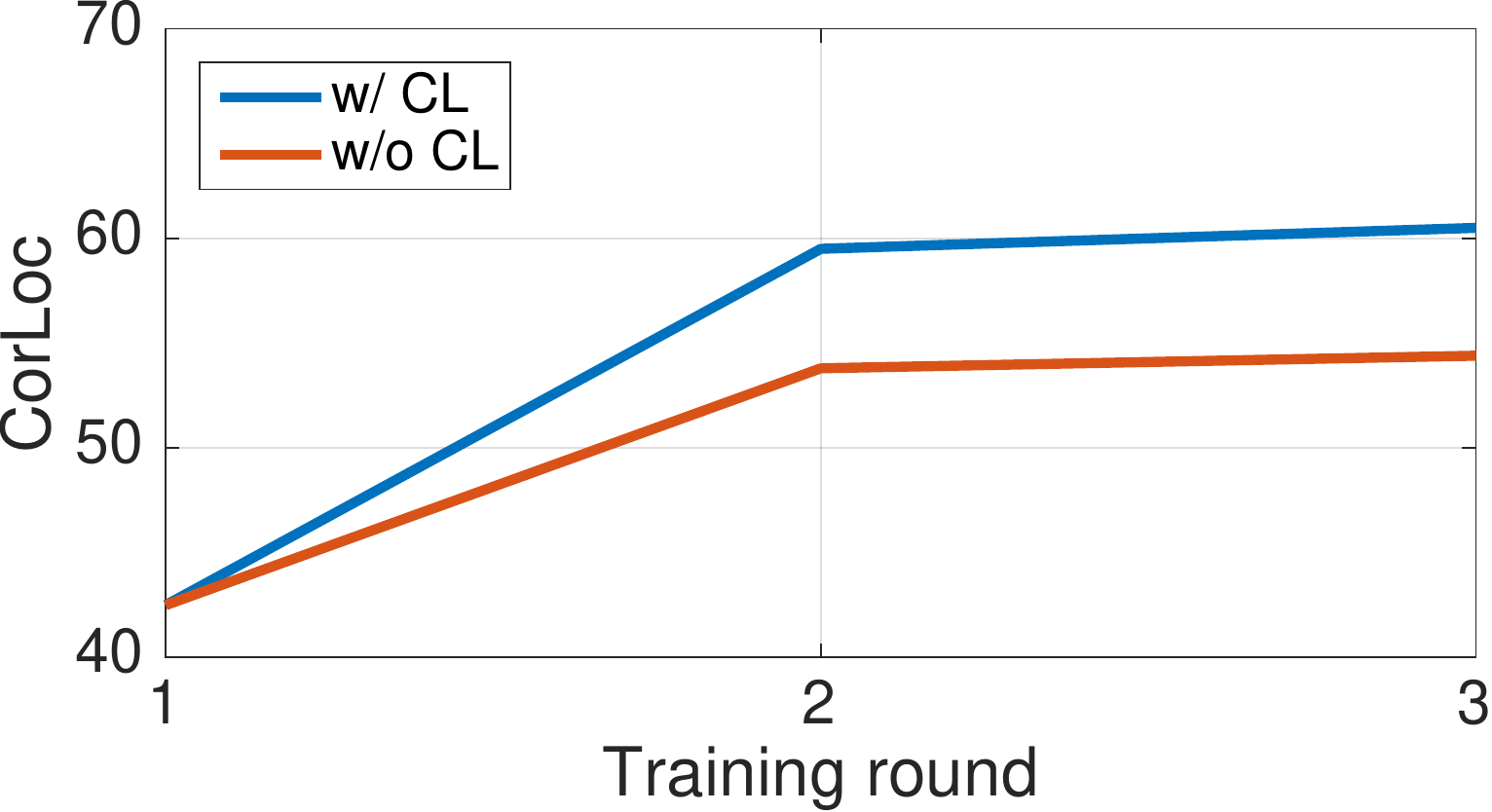}
		\vspace{-0.1cm}
		\caption{Comparison of the CorLoc performance of the Fast R-CNN detector 
			with and without curriculum learning.}\label{fig:corloc_change}
	\end{minipage}
	\hspace{0.1cm}
	\begin{minipage}{0.48\columnwidth}
		\centering
		\includegraphics[height=0.4\columnwidth]{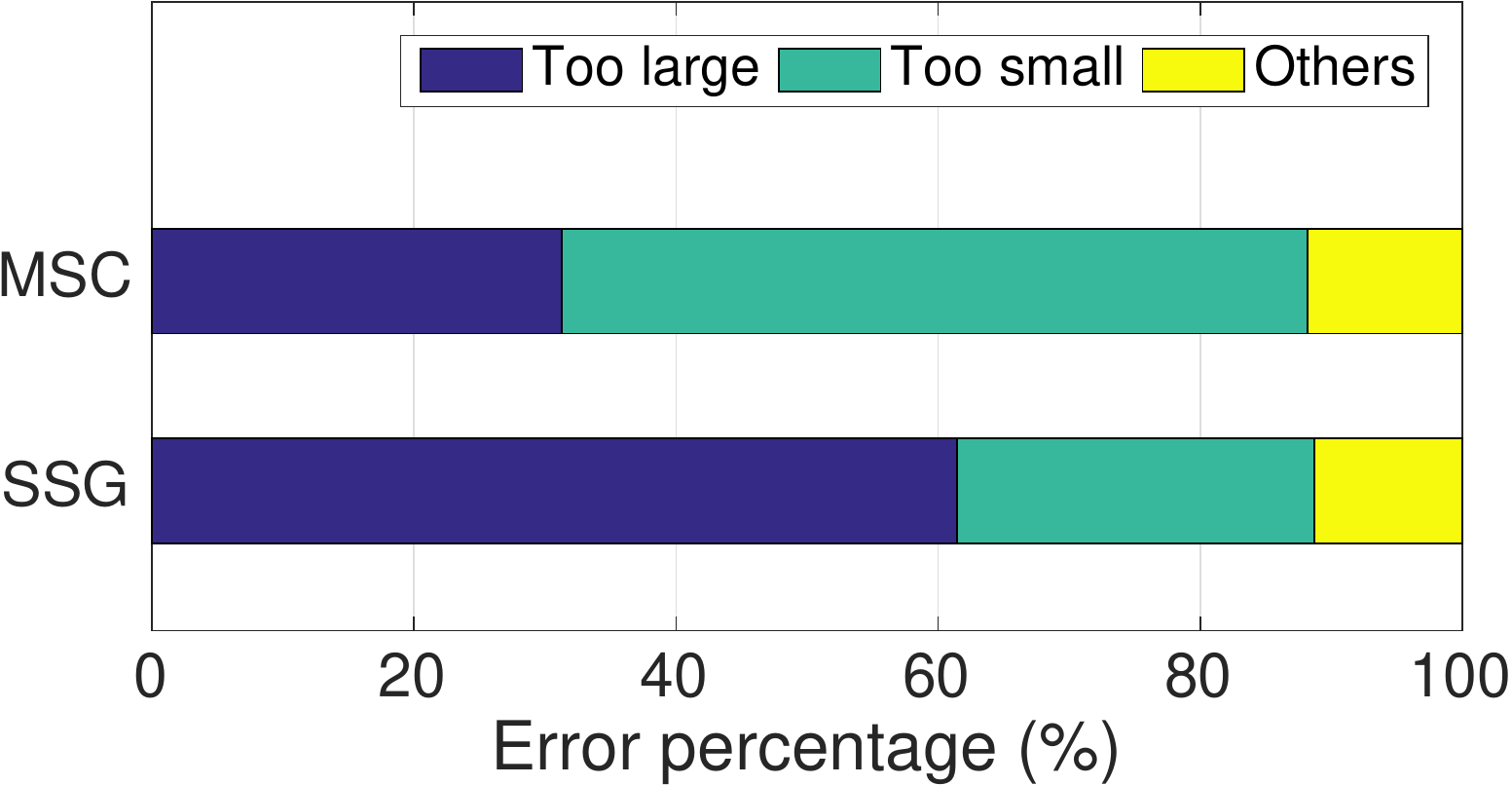}
		\vspace{-0.1cm}
		\caption{Percentages of three error types among the mis-localized objects from the SSG and the MSC.} \label{fig:error_stat}
	\end{minipage}
\end{figure}

\vspace{-0.2cm}
\noindent\textbf{On effectiveness of curriculum learning.}
To prove the effectiveness of curriculum learning, we set up
three baseline Fast R-CNN detectors: (1) trained with pseudo
location labels from the initial MSC detector, with no
segmentation cue used, (2) trained with pseudo location labels from the SSG network, and (3) based on MIL yet without curriculum (where the subset of training examples is selected randomly in each round and the average of $B^{SSG}(c;z)$ and $B^{DET}(c;z)$ is used as pseudo location labels). The ablation study is conducted on the VOC07 dataset. The comparison of the MICL approach and the three baselines are illustrated in Tab.~\ref{tbl:ablation}.
%\todo{I reordered the table columns to be MSC, SSG, MIL, MICL, from the worst to the best, and explained the ablation results. feel free to edit if it doesn't make sense.}.
Adding the segmentation cue into the detector training boosts
the CorLoc by $10.9\%$ and mAP by $4.4\%$ from the initial MSC
detector. In parallel, MIL without curriculum gives us
even slightly larger improvement of $11.9\%$ CorLoc and
$6.3\%$ mAP. By introducing curriculum learning (CL), we obtain an
total improvement of $18.0\%$ in the CorLoc and $10.6\%$ in the mAP. We also compare the CorLoc change during training with and without CL in Fig.~\ref{fig:corloc_change}.
%%%%%%%%%%%%%%%%%%%%%%%%%%%%%%%%%%%%%%%%%%%%%%%%%%%%%%%%%%%%%%%%
\begin{table}[!hbtp]
	\fontsize{8}{9}\selectfont
	\parbox{.48\linewidth}{
		\begin{center}
			\begin{tabular}{|l|c|c|c|c|}
				\hline
				&MSC &SSG &MIL & MICL \\
				\hline\hline
				CorLoc&42.5 & 53.4  &54.4 &\textbf{60.5}\\
				mAP &32.0 &36.4 &38.3 &\textbf{42.6} \\
				\hline
			\end{tabular}
		\end{center}
		\caption{Performance comparison of the three Fast R-CNN baselines and the one with the proposed multiple instance curriculum learning paradigm.}\label{tbl:ablation}
	}
	\hspace{0.1in}
	\parbox{.49\linewidth}{
		\begin{center}
			\begin{tabular}{|l|c|c|c|}
				\hline
				&Subset &All (SSG) &All (MSC) \\
				\hline\hline
				CorLoc & \textbf{72.9} & 53.2 &42.4 \\
				mAP & \textbf{38.0} & 36.3 &32.0\\
				\hline
			\end{tabular}
		\end{center}
		\caption{Comparison of the CorLoc on the selected 
			training subset versus on the whole set and mAP on the \textit{test} set achieved by the correspondingly trained detectors.}\label{tbl:subset_vs_all}
	}
\end{table}
%%%%%%%%%%%%%%%%%%%%%%%%%%%%%%%%%%%%%%%%%%%%%%%%%%%%%%%%%%%%%%%%

\noindent\textbf{On easy example selection criterion.} One reason that curriculum learning is effective is that the pseudo location labels on the selected subset are more reliable than the average, as shown in Tab.~\ref{tbl:subset_vs_all}. This can be explained by the different preferences of the SSG network and the MSC detector when they localize objects: the SSG tends to group close instances from the same category due to the lack of instance information, resulting boxes larger than objects; the MSC detector, however, may focus on the most discriminative regions, usually smaller than the true objects. This intuition is confirmed by Fig.~\ref{fig:error_stat}, where we analyze the localization errors from the SSG and the MSC. Among objects that are mis-localized, errors can be classified into three categories: too large, too small and others\footnote{We provide details about how mis-localized objects are categorized in the supplementary materials.}. It is clear that the SSG tends to generate bounding boxes larger than the target while the MSC prefers smaller ones. Thus, if the results from the SSG and the MSC are consistent, the bounding box is neither too small nor too large, indicating reliable locations.  We also train a Fast R-CNN detector on the easy subset only and compare with the detectors trained on the whole set with pseudo locations from SSG and MSC, respectively. As shown in Tab.~\ref{tbl:subset_vs_all}, the one trained with easy examples achieves 1.7\% higher in mAP than the second best, even though it sees fewer examples, which indicates the importance of the quality of pseudo location labels.
%%%%%%%%%%%%%%%%%%%%%%%%%%%%%%%%%%%%%%%%%%%%%%%%%%%%%%%%%%%%%%%%
\vspace{-0.2cm}

%%%%%%%%%%%%%%%%%%%%%%%%%%%%%%%%%%%%%%%%%%%%%%%%%%%%%%%%%%%%%%%%	
\vspace{-0.7cm}
\section{Conclusion}\label{sec:conclusion}\vspace{-0.2cm}
In this work, we proposed an MICL method with a segmentation network injected to overcome the challenge that detectors often focus on the most discriminative regions when trained without manually labeled tight object bounding boxes. In the proposed MICL approach, where the segmentation-guided MIL is organized on a curriculum, the detector is trained to learn the extent of full objects from easy to hard examples and the easiness is determined automatically by measuring the consistency between the results from the current detector and the segmenter. The benefits from the segmentation network and the power of the easy-to-hard curriculum learning paradigm are demonstrated by extensive experimental results.

\vspace{-0.3cm}
\subsection*{Acknowledgements}\vspace{-0.2cm}
This research was partially supported by Ittiam Systems. We thank Yaguang Li, Junting Zhang and Heming Zhang for helpful discussions and proofreading.

\bibliography{bmvc_final}
\end{document}